\pdfoutput=1

\documentclass[11pt]{article}

\usepackage{acl}

\usepackage{times}
\usepackage{latexsym}
\usepackage{amsmath}
\usepackage[T1]{fontenc}

\usepackage[utf8]{inputenc}

\usepackage{microtype}

\usepackage{inconsolata}
\usepackage{times}
\usepackage{latexsym}
\usepackage{colortbl}
\usepackage{graphics}
\usepackage{graphicx}
\usepackage{multirow}
\usepackage{amssymb}
\usepackage{mathrsfs}
\usepackage{amsmath}
\usepackage{bm}
\usepackage[cmintegrals]{newtxmath}
\usepackage{subfigure}
\usepackage{caption}
\usepackage{makecell}
\usepackage{threeparttable}
\usepackage{booktabs}
\usepackage{tabularx}
\usepackage{array}
\usepackage{makecell}
\usepackage{booktabs}
\usepackage{pifont}

\newcommand{\fone}{{F_1}}
\def \redcell#1 {\cellcolor[HTML]{FDDCCE}{#1}}

\def \green#1{\textcolor[HTML]{16aa09}{#1}}

\usepackage[T1]{fontenc}

\usepackage[utf8]{inputenc}

\usepackage{microtype}

\usepackage{latexsym}
\usepackage{graphics}
\usepackage{graphicx}
\usepackage{color}
\usepackage{colortbl}
\usepackage{amssymb}
\usepackage{amsmath}
\usepackage{bbm}
\usepackage{booktabs}

\usepackage[T1]{fontenc}

\usepackage[utf8]{inputenc}

\usepackage{microtype}
\usepackage{multirow}
\usepackage{footmisc}
\usepackage{soul}
\usepackage{arydshln}

\usepackage{algpseudocode}
\usepackage{algorithm}
\usepackage{hyperref}
\usepackage{xspace}

\usepackage{bigstrut}

\title{PSSAT: A Perturbed Semantic Structure Awareness Transferring Method for Perturbation-Robust Slot Filling}

\author{Guanting Dong$^{1}$\thanks{\ \ The first four authors contribute equally. Weiran Xu is the corresponding author. Email: dongguanting@bupt.edu.cn},
Daichi Guo$^{1*}$,
Liwen Wang$^{1*}$,
Xuefeng Li$^{1*}$,
Zechen Wang$^{1}$,    \\
\textbf{Chen Zeng$^{1}$,
Keqing He$^{2}$,
Jinzheng Zhao$^{3}$,
Hao Lei$^{1}$,
Xinyue Cui$^{1}$,}    \\
\textbf{Yi Huang$^{4}$,
Junlan Feng$^{4}$,
Weiran Xu$^{1*}$}\\ 
$^1$Beijing University of Posts and Telecommunications, Beijing, China\\
$^{2}$Meituan Group, Beijing, China\\
$^{3}$University of Surrey, UK\\
$^{4}$China Mobile Research Institute\\
}

\begin{document}
\maketitle
\begin{abstract}
Most existing slot filling models tend to memorize inherent patterns of entities and corresponding contexts from training data. However, these models can lead to system failure or undesirable outputs when being exposed to spoken language perturbation or variation in practice. We propose a perturbed semantic structure awareness transferring method for training perturbation-robust slot filling models. Specifically, we introduce two MLM-based training strategies to respectively learn contextual semantic structure and word distribution from unsupervised language perturbation corpus. Then, we transfer semantic knowledge learned from upstream training procedure into the original samples and filter generated data by consistency processing. These procedures aim to enhance the robustness of slot filling models. Experimental results show that our method consistently outperforms the previous basic methods and gains strong generalization while preventing the model from memorizing inherent patterns of entities and contexts.

\end{abstract}

\section{Introduction}

The slot filling (SF) task in the goal-oriented dialog system aims to identify task-related slot types in certain domains for understanding user utterances. Traditional supervised slot filling models and sequence labeling methods \cite{liu2015recurrent,Liu2016AttentionBasedRN,goo2018slot,niu2019novel,he2020multi,he2020learning,Wang2022InstructionNERAM} have shown remarkable performance. However, these models tend to memorize inherent patterns of entities and contexts    \cite{wang2022miner,lin2021rockner}. Faced with uncertainty and diversity of human language expression, the perturbation of entities and contexts will lead to a decrease in the generalization ability of the SF model, which hinders its further application in practical dialog scenarios.

\begin{figure}[t]
\centering

\resizebox{.47\textwidth}{!}{\includegraphics{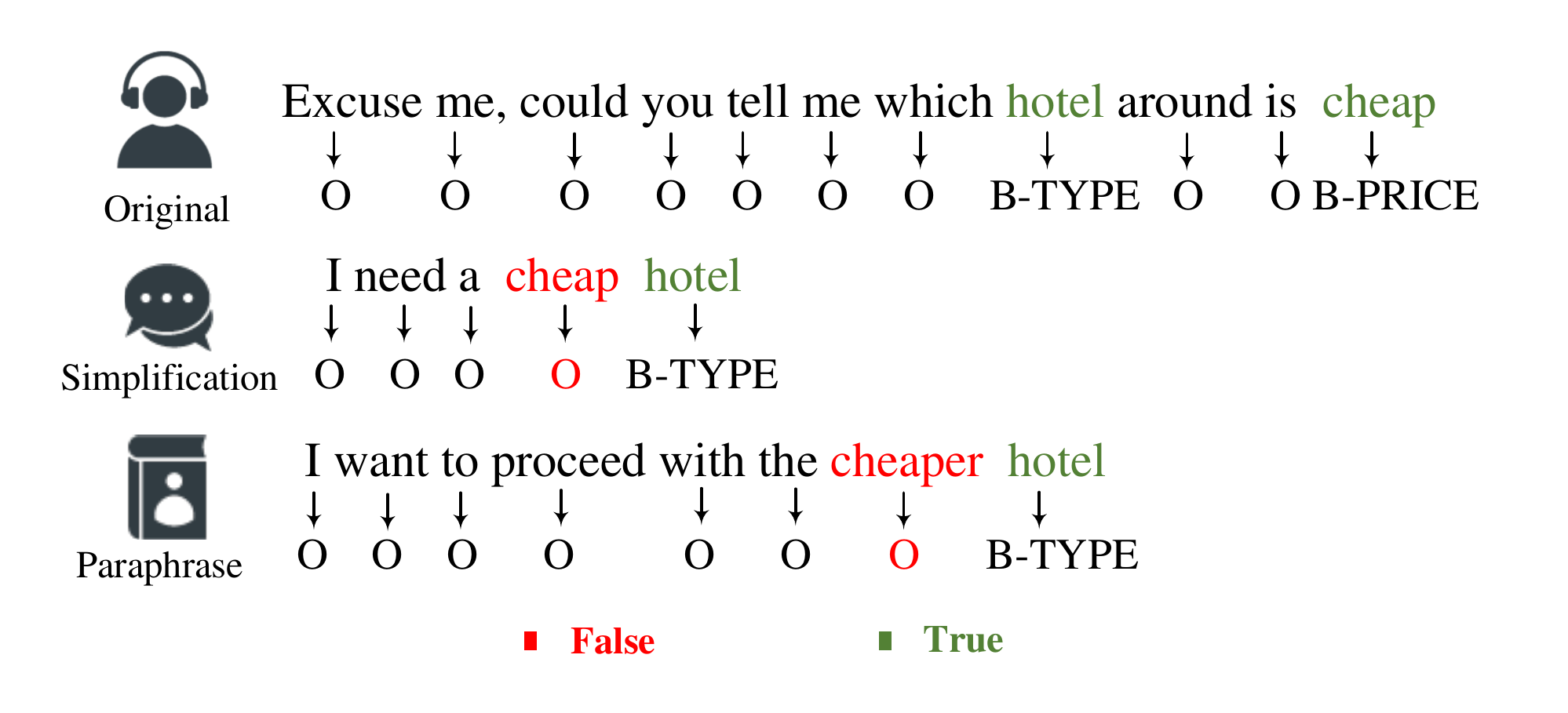}}
\vspace{-0.5cm} 
\caption{The impact of diverse spoken language perturbations on the slot filling\label{fig:intro} system in real scenarios.}

\vspace{-0.8cm} 
\end{figure}

Due to the variety of expression habits, users may not interact with the dialogue system abiding by a rigid input mode in real dialog scenarios. Instead, the expression styles of users would be of high lexical and syntactic diversity while users express their intentions. An interesting finding is that, every expression retains the key semantic information of the sentence to ensure consistency of the intention, but it inevitably damages the semantic structure of the context. As shown in Figure \ref{fig:intro}, the original sentence comes from training data, while the other two sentences are real queries of users with different language habits. Firstly, paraphrase and simplification perturb the contextual semantic structure of the original sentence to various degrees. Secondly, some slot entities also suffer from word perturbations. However, they all retain price-related information to express the same intention. We refer to the above two perturbations collectively as Spoken Language Perturbation. The previous slot filling model, which tends to memorize entity patterns, has a significantly reduced generalization ability when faced with these situations. Therefore, it is necessary to train a robust slot filling model against perturbations in practical application.

\begin{figure*}[ht]
\centering
\resizebox{1\textwidth}{!}{\includegraphics{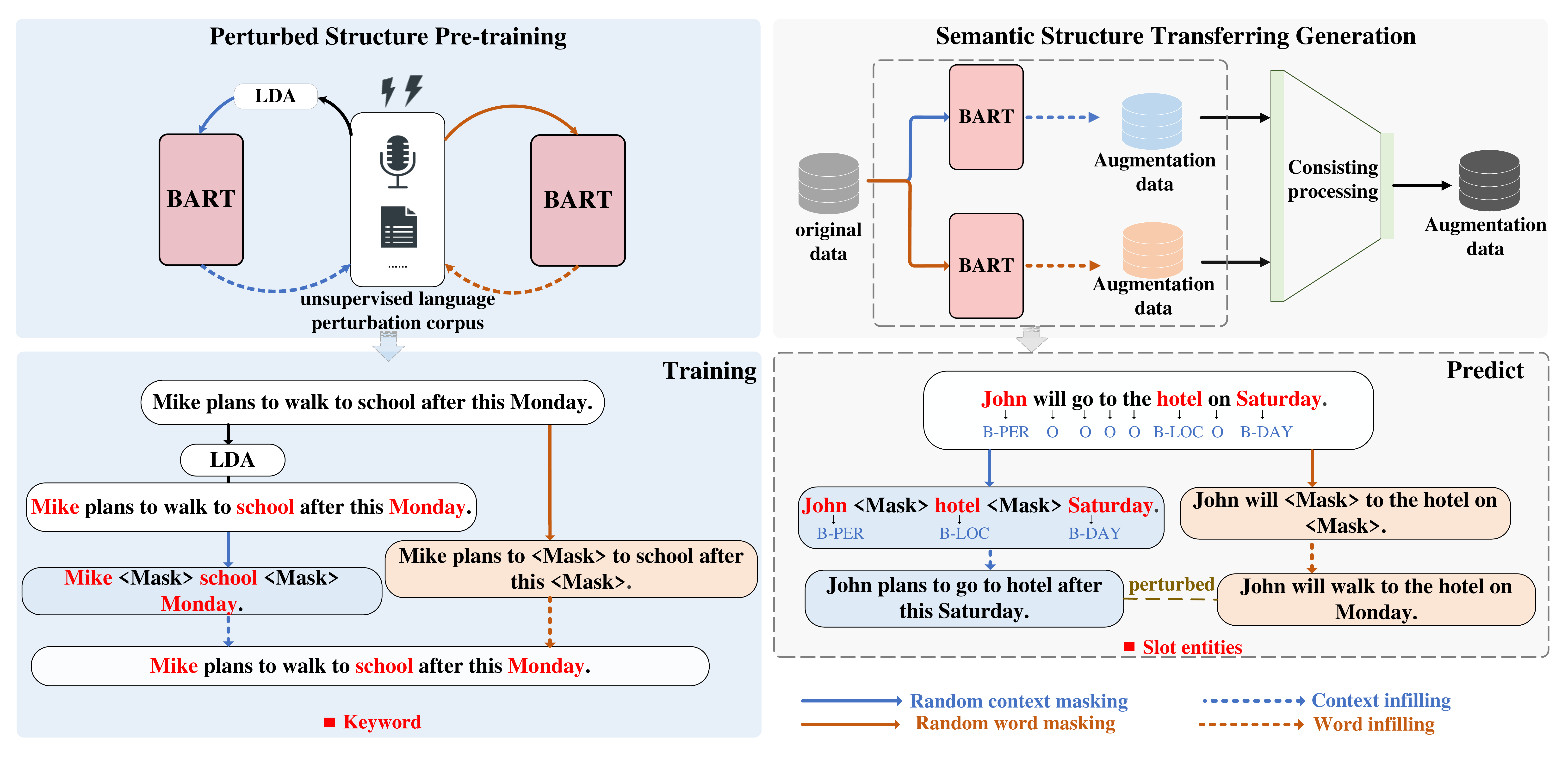}}
\vspace{-0.8cm}
    \caption{The overall architecture of the PSSAT framework. Two dotted boxes show the specific processes of the MLM-based strategies at pre-training and transferring generation stage, respectively.}
 
\label{fig:method}
\vspace{-0.3cm} 
\end{figure*}


Recently, improving the robustness of NLP systems against input perturbations has attracted increasing attention. Most existing studies   \cite{wu2021bridging,moradi2021evaluating,gui2021textflint} that explored the robustness problem are only about rule-based synthetic datasets, which have certain limitations. Further, \citet{namysl2020nat} focused on the robustness of the NER model against Optical Character Recognition (OCR) disturbance and misspellings. However, real-world dialogue systems face more diverse perturbations due to frequent interactions with users. \citet{liu2020robustness} proposed Language understanding augmentation, which contains four data augmentation methods, to simulate natural perturbations. Nevertheless, each method is designed for a specific perturbation, which cannot generalize for other unknown perturbations.

To solve the above issues, in this paper, we propose a \textbf{P}erturbed \textbf{S}emantic \textbf{S}tructure \textbf{A}wareness \textbf{T}ransferring method (\textbf{PSSAT}). It can generate augmented data based on human diversity expressions. In fact, it is not difficult to obtain unsupervised corpora containing spoken language perturbations in real-world scenarios (e.g. social media). Therefore, we extract the texts from two multi-modal datasets \cite{zhang2018adaptive,lu2018visual} and construct an unsupervised language perturbation corpus, which helps the model learn the semantic structure of perturbed data. To be specific, we introduce a perturbed structure pre-training stage, which guides the model to directly learn contextual semantic structure and words distribution from unsupervised language perturbation corpus through two different MLM-based training strategies, respectively. To better eliminate the distribution gap between upstream and downstream data, we design a \emph{Semantic Structure Transferring Generation} stage to transfer the upstream learned semantic structure knowledge to downstream original training samples. By doing so, the generated augmented samples are more in line with the spoken language perturbation. However, as there are mixed perturbations existed in upstream corpus, the model may generate some low-quality samples. To alleviate this problem, we introduce \emph{Consistency Processing} to filter generated samples. 

Our contributions are three-fold: (1) To the best of our knowledge, this is the first work to investigate spoken language perturbation of slot filling tasks and validate the vulnerability of existing rule-based methods in the condition of diverse language expressions. (2) We propose a perturbed semantic structure awareness transferring  method, which transfers the learned contextual semantic structure and word distribution into the original samples through the MLM-based method. (3) Experiments demonstrate that our method outperforms all baseline methods and gains strong generalization while preventing the model from memorizing inherent patterns of entities and contexts.

\vspace{-0.3cm}

\section{Methodology}

\subsection{Problem Definition}
Given a tokenized utterance $X=\{x_1, x_2, \dots, x_N\}$ and its corresponding BIO format label $Y=\{y_1, y_2, \dots, y_N\}$,
we formulate the spoken language perturbation process in the real scenario as $X'=\mathcal{P}_x(X),   Y'=\mathcal{P}_y(Y)$
such that $X' \ne X$ but $Y'$ may be identical with $Y$ or not. The perturbation-robust slot filling
requires the model to be tested on the perturbed test dataset $\{(X',Y')\}$ but with no access to the spoken language perturbation process $\mathcal{P}(\cdot)$ 
or perturbed data during the training phase.


\subsection{Perturbed Structure Pre-training}






The perturbed structure pre-training stage guides the model to learn the semantic structure from realistic perturbed data. We carefully collected several spoken language perturbation datasets to build an unsupervised language perturbation corpus\footnote{More details about the construction process of the perturbation corpus can be found at section \ref{ref:B}}.  
 Inspired by the key idea of masked language model (MLM) \cite{devlin2018bert}, which randomly replaces a few tokens in a sentence with the special token \emph{[MASK]} and recovers the original tokens by a neural network, we introduce two augmentation strategies, as shown in Figure \ref{fig:method}.

Random Word Masking (RWM): words are randomly selected for masking and infilling to simulate the word perturbation, which guides the model to learn word distribution from real perturbed data.

Random Context Masking (RCM): we filter out the keywords of each sentence through Latent Dirichlet Allocation (LDA) \cite{blei2003latent} to keep the key information of the sentence. For non-keyword parts, we regard them as context spans of each sentence and conduct random masking and infilling. In this way, the model learns the contextual semantic structure from realistic perturbed data. Unlike word infilling, context infilling can generate multiple tokens for each \emph{[MASK]} position.

\vspace{-0.3cm}

\subsection{Semantic Structure Transferring Generation}
The Semantic Structure Transferring Generation stage aims to transfer learned contextual semantic structure and word distribution from upstream pre-trained model to downstream training samples. As shown in Figure \ref{fig:method}, pre-trained models are separately loaded to conduct RWM and RCM. A slight difference from the pre-training stage is that slot entities are filtered out as keywords. 
It is worth noting that augmented data generated by two strategies explicitly contain diverse human expressions, which are learned from perturbed structure pre-training. Besides, we also generate coarse labels for two kinds of augmented data based on rules. Specifically, we label the infilling tokens as \emph{O} while maintaining labels of other tokens. The case study (See Appendix \ref{ref:H}) shows that samples generated by semantic structure transferring generation can not only better fit spoken language perturbation, but also be more in line with human language diversity than those generated by rule-based methods.

\begin{figure}[t]
\centering

\resizebox{.47\textwidth}{!}{\includegraphics{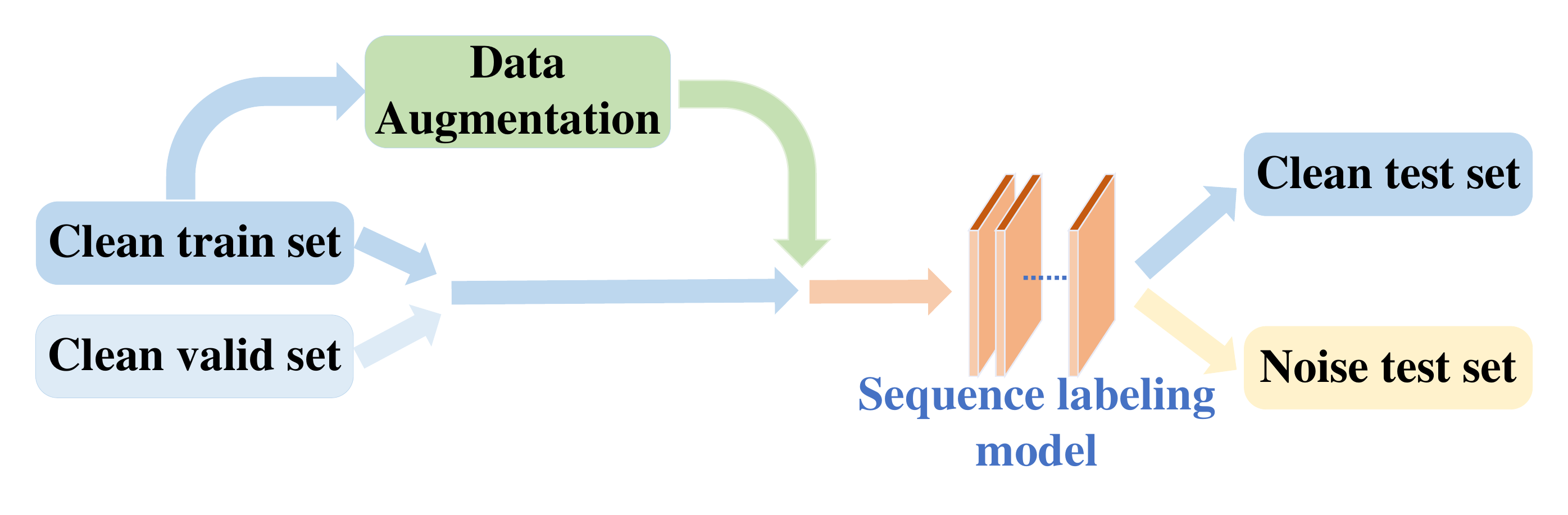}}
\caption{The process of downstream perturbation-robust slot filling task.}
\label{fig:process}
\vspace{-0.5cm} 
\end{figure}

\textbf{Consistency Processing}
Due to mixed perturbations in the upstream corpus, it is necessary to design a consistency processing to filter low-quality samples. Specifically, we train a tagging model with original training data and augmented samples. Then the model is used to predict labels for each augmented sentence. The labels which are consistent with the coarse labels and original labels are kept. The augmented samples filtered by consistency processing are mixed and input to the main task as the final augmented data.

\textbf{Training and Inference}
As shown in figure \ref{fig:process}, during the training stage, we first perform perturbed structure pre-training on the unsupervised language perturbation corpus to learn the contextual semantic structure and word distribution of perturbed data. We use the pre-trained model to obtain augmented data for the clean training dataset, and use all samples to train a perturbation-aware sequence labeling model. During the testing stage, we test the sequence labeling model on both clean and perturbed datasets.



\begin{table*}[t]
  \centering
\resizebox{140mm}{30mm}{
  \begin{tabular}{lcccccccccc}
    \toprule
    \textbf{Methods}&\textbf{Clean}&\textbf{Homophone}&\textbf{Paraphrase}&\textbf{Verbose}&\textbf{Simplification}&\textbf{Overall} \\
    
    \midrule
    none &
    95.8&
    81.5 (-14.3)&
    87.5 (-8.3)&
    81.6 (-14.2)&
    85.3 (-10.5)&
    84 (-11.8)\\
    \midrule
    
    Char-Random &
    96.0 (0.2)&
    84.1 (18.2\%)&
    87.6 (1.2\%)&
    83.2 (11.3\%)&
    88.1 (26.7\%)&
    85.8 (14.4\%)\\
    
    Word-Del &
    95.9 (0.1)&
    83.2 (11.9\%)&
    89.3 (21.7\%)&
    82.6 (7.0\%)&
    87.5 (21.0\%)&
    85.7 (15.4\%)\\
    
    Syn-Sub &
    96.1 (0.3)&
    83.5 (14.0\%)&
    89.3 (21.7\%)&
    82.2 (4.2\%)&
    86.8 (14.3\%)&
    85.5 (13.6\%)\\
    
    Word-Insert &
    95.8 (0.0)&
    \textcolor{red}{81.2 (-2.1\%)}&
    88.2 (8.4\%)&
    \textcolor{red}{81.3 (-2.1\%)}&
    \textcolor{red}{86.2 (8.6\%)}&
    \textcolor{red}{84.2 (3.2\%)}\\    
    
    Hom-Sub &
    96.0 (0.2)&
    83.7 (15.4\%)&
    89.3 (21.7\%)&
    82.3 (4.9\%)&
    87.7 (22.9\%)&
    85.8 (16.3\%)\\   
    
    NAT($\mathcal{L}_{\textit{aug}}$) &
    96.0 (0.2)&
    \underline{84.3 (19.6\%)}&
    87.7 (2.4\%)&
    82.8 (8.5\%)&
    87.3 (19.0\%)&
    85.5 (12.4\%)\\

    NAT($\mathcal{L}_{\textit{stabil}}$) &
    96.0 (0.2)&
    83.9 (16.8\%)&
    \textcolor{red}{87.4 (-1.2\%)}&
    83.0 (9.9\%)&
    87.3 (19.0\%)&
    85.4 (11.1\%)\\  
     \midrule
    
    
    PSSAT  &
    96.2 (0.4)&
   \textbf{ 84.6 (21.7\%)}&
    \textbf{90.1 (31.3\%)}&
   \textbf{84.0 (16.9\%)}&
    \textbf{89.3 (38.1\%)}&
    \textbf{87.0 (27.0\%)}\\
    
    \quad$-$ RCM &
    96.2 (0.4)&
    83.8 (16.1\%)&
    89.6 (25.3\%)&
    83.5 (13.4\%)&
    87.4 (20.0\%)&
    86.1 (18.7\%) \\

    \quad$-$ RWM&
    96.3 (0.5)&
    83.3 (12.6\%)&
    89.9 (28.9\%)&
    \underline{83.8 (15.5\%)}&
    \underline{88.9 (34.3\%)}&
    \underline{86.5 (22.8\%)}\\
    
    \quad$-$ CP&
    96.3 (0.5)&
    84.0 (17.5\%)&
    \underline{90.0 (30.1\%)}&
    83.4 (12.7\%)&
    88.3 (28.6\%)&
    86.4 (22.2\%)\\
    
    \quad$-$ Pre-training&
    95.9 (0.1)&
    83.1 (11.2\%)&
    89.4 (22.9\%)&
    83.0 (9.9\%)&
    86.9 (15.2\%)&
    85.6 (14.8\%)\\
    \bottomrule

  \end{tabular}}
  \caption{The performance (F1 score) of the PSSAT on RADDLE. For cells in \textit{Baseline} row and \textit{Clean test} column, the numbers in the parenthesis indicate the change of F1 score. For other cells, the numbers in the parenthesis indicate ${p_r}$. In Overall column, we calculate the average F1 and ${p_r}$ of the four Spoken language perturbations, respectively. Both the best and the worst are marked, "$-$" denotes the model performance without a specific module. RWM, RCM, CP denotes Random Word Masking, Random Context Masking and Consistency Processing.}
  \label{tab:noise sf}
  \vspace{-0.3cm}
  
\end{table*}

\begin{table}[htbp]
    \centering
    \tiny
    \renewcommand{\arraystretch}{1.5}
    \setlength{\tabcolsep}{2pt}
   \resizebox{76mm}{11mm}{
    \begin{tabular}{ccccccccc}
        \hline
        \textbf{Method}   &  \textbf{Hom+App} & \textbf{Hom+Con} & \textbf{Con+App} & \textbf{Hom+Con+App}  \\

        \hline
        Baseline (LSTM)  & 47.9 (-46.0) & 54.2 (-39.7) &  73.4 (-20.5) &  45.7 (-48.2) \\

        best baseline  & 53.6 (12.4\%) & 61.1 (17.4\%) &  71.6 (-8.8\%) &  47.2 (3.1\%) \\ 
        
        \hline
        
        PSSAT   & 59.6 (25.4\%) & 61.8 (19.1\%) & 78.3 (23.9\%) & 53.9 (17.0\%) & \\    
        
        \hline

    \end{tabular}}
    \vspace{-0.2cm}
    \caption{The performance of the best baseline and PSSAT on mixed perturbations. }
    \label{tab:snips main}
    \vspace{-0.6cm}
\end{table}

\vspace{-0.3cm}
\section{Experiment}

\subsection{Dataset}
\textbf{RADDLE} \cite{peng2020raddle} is a crowd-sourced diagnostic dataset to cover a broad range of real-world perturbations to study the robustness of end-to-end dialog system. We extract four kinds of realistic perturbed data from RADDLE and construct the slot filling dataset. In particular, the original dataset of the evaluation set in RADDLE is extracted from MultiWOZ \cite{lu-etal-2021-slot}. To introduce sufficient perturbed data for evaluating the model robustness against multiple perturbations, we extracted the clean user utterances and four kinds of perturbed utterances (Homophone, Simplify, Verbose and Paraphrase) from RADDLE. To be specific, \textbf{Homophone} perturbation comes from input text errors caused by recognition and synthesis errors. \textbf{Simplification} is generated by concise-word expression.  On the contrary, \textbf{Verbose} refers to redundant expression. \textbf{Paraphrase} noise widely exists in our dataset, where users restate texts in different ways of expression according to their personal speaking habits. The training dataset consists of 61,117 clean data from four domains. We randomly select 5,000 data as the validation set. Our compared baselines and implementation details can be found in Section \ref{ref:C} and \ref{ref:D}.


\subsection{Unsupervised Language Perturbation Corpus} 
\label{ref:B}
In our perturbed structure pre-training stage, we employ two multi-modal datasets: Twitter-2015 \cite{zhang2018adaptive}, Twitter-2017 \cite{lu2018visual}. We only extract the corpus part and delete the useless details in sentences such as emoji and URL. We consider that the data on social media contains the real diversity of human expressions, and it is beneficial for the downstream generation to learn the knowledge of diverse human expressions in the pre-training stage.


\subsection{Evaluation Metrics}

We use $\fone$ score to measure the performance of the model. $\fone^{c}$, $\fone^{p}$ denote the performance on the clean and perturbed test set respectively. On this basis, we define (1) as Perturbation Recovery Rate (${P_r}$) of a given perturbation-robust method $m$:
\begin{equation}
\begin{aligned}
{P_r} = \frac{{F_1}_m^p-{F_1}_{\rm baseline}^p}{{F_1}_{\rm baseline}^c-{F_1}_{\rm baseline}^p}
\end{aligned}
\end{equation}
 ${P_r}$ indicates the improvement in performance of the model using the robust approach over the baseline model on the perturbed test set, as a percentage of the performance degradation of the baseline model due to the introduction of perturbation.


        


\subsection{Implementation Details}
\label{ref:C}
For the upstream work, our model \textbf{PSSAT} is based on BART \cite{lewis2019bart}, which is provided by the Huggingface Transformers. The reason for choosing BART is that the pre-training tasks of BART include token masking and text filling, which is consistent with our PSSAT task. We set the batch size of BART to 8 and the pre-training takes an average of one hour for 10 epochs. The corresponding learning rates are set to 1e-5.

For the downstream work, we use two settings for perturbation-robust slot filling, Glove-Bi-LSTM and BERT-Bi-LSTM. Glove-6B-300d, char embedding and BERT-large-uncased are applied as the embedding layer. We take Bi-LSTM as the mainly analyzed model. The hidden size of Bi-LSTM is set to 128 and the dropout rate is set to 0.2. The transform probability \textit{p} is set to 0.3. For all the experiments, we train and test our model on the 2080Ti GPU. It takes an average of 1.5 hours to run with 12 epochs on the training dataset.

All experiments are repeated three times with different random seeds under the same settings. All the models are implemented with PyTorch \cite{paszke2019pytorch}.

\vspace{-0.15cm}
\subsection{Main Results} 

Table \ref{tab:noise sf} shows the main results of PSSAT compared to different baselines on the language perturbation dataset. The overall result of our PSSAT greatly outperforms the baseline by 27.0\%. Especially, the ${P_r}$ of paraphrase and simplification is about 40\%, which is a remarkable enhancement. What's more, our method is not designed for any specific perturbation, but achieves the best results for various perturbations, which proves that our model not only improves the performance significantly, but generalizes better. 

\textbf{Ablation Studies.} To better prove the effectiveness of the pre-training stage, we conduct ablation experiments. Table \ref{tab:noise sf} illustrates the results that the model without RWM performs better than that without RCM, which shows that the change of context makes the semantic change more drastic. Meanwhile, all of RWM, RCM, CP and PSSAT without pre-training  have a performance drop, which suggests that every part of  design is necessary.


\vspace{-0.15cm}

\subsection{Mixed Perturbations Experiment}
In real dialogue scenarios, mixed perturbations often appear in one input utterance at the
same time. To verify the effectiveness of our method in more realistic scenarios,
based on SNIPS \cite{coucke2018snips}, we utilize TextFlint\footnote{http://textflint.io/
} \cite{gui2021textflint} to introduce Homophone(Hom), Appendirr(App), ConcatSent(Con) and construct a mixed perturbations evaluation dataset \footnote{We conducted single perturbation experiment on SNIPS. The results can be found in Appedix \ref{ref:F}}. As shown in Table \ref{tab:snips main}
, the ${P_r}$ of our PSSAT is over 20\% against three different kinds of  two-level perturbations, which far exceeds the best baseline (Hom-Sub). The model maintains an almost 17\% ${P_r}$ even with the joint disturbances from three-level perturbations, which shows the effectiveness and stability of our methods in real scenarios.

\vspace{-0.15cm}
\subsection{Error Analysis}

We randomly selected 500 samples from all outputs and manually checking the error outputs for error analysis. Table \ref{tab:error} investigates 5 error types the model has made on the RADDLE. It can be seen that the number of PSSAT error outputs is less than the baseline in each category. Table \ref{tab:error case} illustrates cases of each error type. Both the baseline model and PSSAT can correctly label clean text, but only PSSAT can correctly label texts with perturbation. After comprehensive analysis, the result shows that rote memorization of entity mention and contextual perturbation accounts for a large portion of the errors. Compared to the baseline, PSSAT can alleviate the problem of memorizing inherent patterns of entities and contexts.


\begin{table}[t]
  \centering
  \small
  \resizebox{72mm}{15mm}{
    \begin{tabular}{ccccr}
    \hline
    \multirow{2}[2]{*}{Error Type} & \multicolumn{2}{c}{Baseline} & \multicolumn{2}{c}{PSSAT} \bigstrut[t]\\
          & Num   & \%    & Num   & \multicolumn{1}{c}{\%} \bigstrut[b]\\
    \hline
    Entity Location & 12    &  20.0     & 9    & 18.8  \bigstrut[t]\\
    Contextual Perturbation & 16    & 26.7      & 11    &22.8  \\
    Entity Mention & 23    &   38.3    & 19    &39.6  \\
    Others & 9     &    15.0   & 9     & 18.8 \bigstrut[b]\\
    \hline
    Mixed Perturbation & 11    &$-$   & 9     &$-$ \bigstrut\\
    \hline
    \end{tabular}%
    }
    \caption{Error analysis on RADDLE. }
  \label{tab:error}%

\end{table}%

\label{ref:G}

\begin{table}[t]
  \centering
    \small
\resizebox{76mm}{35mm}{    
    \begin{tabular}{l|l}
    \hline
    Clean & are there any \textbf{museums} in the \textbf{centre} ? \bigstrut[t]\\
    Verbosity & \multicolumn{1}{p{16.72em}}{could you please search for any \textbf{museums} in the town \textbf{centre} .} \\
    Baseline & O O O O O O B-type O O O \textcolor{red}{O} O\\
    PSSAT  & O O O O O O B-type O O O \textcolor[RGB]{0,119,51}{B-area} O \bigstrut[b]\\
    \hline
    Clean & i 'd like a \textbf{jamaican} restaurant please . \bigstrut[t]\\
    Simplification & find \textbf{jamaican} plz .\\
    Baseline & O \textcolor{red}{B-name} \textcolor{red}{I-name} O\\
    PSSAT  & O \textcolor[RGB]{0,119,51}{B-food} \textcolor[RGB]{0,119,51}{O} O \bigstrut[b]\\
    \hline
    Clean  & i need to leave after \textbf{12:00} .\bigstrut[t]\\
    Homophone & i need to leave after \textbf{twelve} .\\
    Baseline & O O O O O \textcolor{red}{O} O\\
    PSSAT  & O O O O O \textcolor[RGB]{0,119,51}{B-leave} O\bigstrut[b]\\
    \hline
    Clean & i need a booking for \textbf{4} people . \bigstrut[t]\\
    Paraphrase & i need seats for \textbf{4} .\\
    Baseline & O O O O \textcolor{red}{B-time} O\\
    PSSAT  & O O O O \textcolor[RGB]{0,119,51}{B-people} O\bigstrut[b]\\
    \hline
    Clean  & could you tell me which hotel around is \textbf{cheap} ? \bigstrut[t]\\
    Paraphrase & I want to proceed with the \textbf{cheaper} hotel . \\
    Baseline & O O O O O O \textcolor{red}{O} B-type O\\
    PSSAT  & O O O O O O \textcolor[RGB]{0,119,51}{B-price} B-type O\bigstrut[b]\\
    \hline
    \end{tabular}%
    }
        \caption{The error cases. The bold texts are slot entities. Both wrong and correct labels are marked in red and green, respectively.}

  \label{tab:error case}%
\vspace{-0.5cm}
\end{table}%

\vspace{-0.15cm}

\section{Conclusion}
\label{sec:bibtex}
\vspace{-0.15cm}

In this paper, we propose a perturbed semantic structure awareness transferring method for perturbation-robust slot filling task. Specifically, we design the perturbed structure pre-training and the semantic structure transferring generation to transfer the upstream learned semantic structure knowledge to downstream original training samples. Further, we filter low-quality samples through a consistency processing module. Sufficient experiments and error analysis demonstrate the effectiveness and generalization of our methods, and also prove that PSSAT alleviates the problem of memorizing inherent patterns of entities and contexts.

\section*{Acknowledgements}
We thank all anonymous reviewers for their helpful comments and suggestions. This work was partially supported by MoE-CMCC "Artifical Intelligence" Project No. MCM20190701, National Key R\&D Program of China No. 2019YFF0303300 and Subject II No. 2019YFF0303302, DOCOMO Beijing Communications Laboratories Co., Ltd.

\appendix

\bibliography{anthology,custom}

\begin{thebibliography}{}

\bibitem[Agarwal et~al., 2020]{DBLP:journals/corr/abs-2004-04564}
Agarwal, O., Yang, Y., Wallace, B.~C., and Nenkova, A. (2020).
\newblock Interpretability analysis for named entity recognition to understand
  system predictions and how they can improve.
\newblock {\em CoRR}, abs/2004.04564.

\bibitem[Agarwal et~al., 2021]{10.1162/coli_a_00397}
Agarwal, O., Yang, Y., Wallace, B.~C., and Nenkova, A. (2021).
\newblock {Interpretability Analysis for Named Entity Recognition to Understand
  System Predictions and How They Can Improve}.
\newblock {\em Computational Linguistics}, 47(1):117--140.

\bibitem[Aho and Ullman, 1972]{Aho:72}
Aho, A.~V. and Ullman, J.~D. (1972).
\newblock {\em The Theory of Parsing, Translation and Compiling}, volume~1.
\newblock Prentice-Hall, Englewood Cliffs, NJ.

\bibitem[{American Psychological Association}, 1983]{APA:83}
{American Psychological Association} (1983).
\newblock {\em Publications Manual}.
\newblock American Psychological Association, Washington, DC.

\bibitem[Ando and Zhang, 2005]{Ando2005}
Ando, R.~K. and Zhang, T. (2005).
\newblock A framework for learning predictive structures from multiple tasks
  and unlabeled data.
\newblock {\em Journal of Machine Learning Research}, 6:1817--1853.

\bibitem[Andrew and Gao, 2007]{andrew2007scalable}
Andrew, G. and Gao, J. (2007).
\newblock Scalable training of {L1}-regularized log-linear models.
\newblock In {\em Proceedings of the 24th International Conference on Machine
  Learning}, pages 33--40.

\bibitem[Bao et~al., 2021]{bao2021defending}
Bao, R., Wang, J., and Zhao, H. (2021).
\newblock Defending pre-trained language models from adversarial word
  substitutions without performance sacrifice.
\newblock {\em arXiv preprint arXiv:2105.14553}.

\bibitem[Belinkov and Bisk, 2017]{belinkov2017synthetic}
Belinkov, Y. and Bisk, Y. (2017).
\newblock Synthetic and natural noise both break neural machine translation.
\newblock {\em arXiv preprint arXiv:1711.02173}.

\bibitem[Blei et~al., 2003]{blei2003latent}
Blei, D.~M., Ng, A.~Y., and Jordan, M.~I. (2003).
\newblock Latent dirichlet allocation.
\newblock {\em Journal of machine Learning research}, 3(Jan):993--1022.

\bibitem[Brill and Moore, 2000]{brill2000improved}
Brill, E. and Moore, R.~C. (2000).
\newblock An improved error model for noisy channel spelling correction.
\newblock In {\em Proceedings of the 38th annual meeting of the association for
  computational linguistics}, pages 286--293.

\bibitem[Budzianowski et~al., 2020]{budzianowski2020multiwoz}
Budzianowski, P., Wen, T.-H., Tseng, B.-H., Casanueva, I., Ultes, S., Ramadan,
  O., and Gašić, M. (2020).
\newblock Multiwoz -- a large-scale multi-domain wizard-of-oz dataset for
  task-oriented dialogue modelling.

\bibitem[Chandra et~al., 1981]{Chandra:81}
Chandra, A.~K., Kozen, D.~C., and Stockmeyer, L.~J. (1981).
\newblock Alternation.
\newblock {\em Journal of the Association for Computing Machinery},
  28(1):114--133.

\bibitem[Chen et~al., 2020a]{chen2020seqvat}
Chen, L., Ruan, W., Liu, X., and Lu, J. (2020a).
\newblock Seqvat: Virtual adversarial training for semi-supervised sequence
  labeling.
\newblock In {\em Proceedings of the 58th Annual Meeting of the Association for
  Computational Linguistics}, pages 8801--8811.

\bibitem[Chen et~al., 2020b]{chen-etal-2020-seqvat}
Chen, L., Ruan, W., Liu, X., and Lu, J. (2020b).
\newblock {S}eq{VAT}: Virtual adversarial training for semi-supervised sequence
  labeling.
\newblock In {\em Proceedings of the 58th Annual Meeting of the Association for
  Computational Linguistics}, pages 8801--8811, Online. Association for
  Computational Linguistics.

\bibitem[Chiu and Nichols, 2016]{chiu2016named}
Chiu, J.~P. and Nichols, E. (2016).
\newblock Named entity recognition with bidirectional lstm-cnns.
\newblock {\em Transactions of the Association for Computational Linguistics},
  4:357--370.

\bibitem[Coucke et~al., 2018]{coucke2018snips}
Coucke, A., Saade, A., Ball, A., Bluche, T., Caulier, A., Leroy, D., Doumouro,
  C., Gisselbrecht, T., Caltagirone, F., Lavril, T., Primet, M., and Dureau, J.
  (2018).
\newblock Snips voice platform: an embedded spoken language understanding
  system for private-by-design voice interfaces.

\bibitem[Coulombe, 2018]{coulombe2018text}
Coulombe, C. (2018).
\newblock Text data augmentation made simple by leveraging nlp cloud apis.

\bibitem[Devlin et~al., 2018]{devlin2018bert}
Devlin, J., Chang, M.-W., Lee, K., and Toutanova, K. (2018).
\newblock Bert: Pre-training of deep bidirectional transformers for language
  understanding.
\newblock {\em arXiv preprint arXiv:1810.04805}.

\bibitem[Fang et~al., 2020]{fang2020using}
Fang, A., Filice, S., Limsopatham, N., and Rokhlenko, O. (2020).
\newblock Using phoneme representations to build predictive models robust to
  asr errors.
\newblock In {\em Proceedings of the 43rd International ACM SIGIR Conference on
  Research and Development in Information Retrieval}, pages 699--708.

\bibitem[Feng et~al., 2021]{feng2021asr}
Feng, L., Yu, J., Cai, D., Liu, S., Zheng, H., and Wang, Y. (2021).
\newblock Asr-glue: A new multi-task benchmark for asr-robust natural language
  understanding.
\newblock {\em arXiv preprint arXiv:2108.13048}.

\bibitem[Fu et~al., 2020]{NERcase2020}
Fu, J., Liu, P., and Zhang, Q. (2020).
\newblock Rethinking generalization of neural models: A named entity
  recognition case study.
\newblock {\em Proceedings of the AAAI Conference on Artificial Intelligence},
  34(05):7732–7739.

\bibitem[Gardner et~al., 2018]{gardner-etal-2018-writing}
Gardner, M., Neumann, M., Grus, J., and Lourie, N. (2018).
\newblock Writing code for {NLP} research.
\newblock In {\em Proceedings of the 2018 Conference on Empirical Methods in
  Natural Language Processing: Tutorial Abstracts}, Melbourne, Australia.
  Association for Computational Linguistics.

\bibitem[Goo et~al., 2018a]{goo-etal-2018-slot}
Goo, C.-W., Gao, G., Hsu, Y.-K., Huo, C.-L., Chen, T.-C., Hsu, K.-W., and Chen,
  Y.-N. (2018a).
\newblock Slot-gated modeling for joint slot filling and intent prediction.
\newblock In {\em Proceedings of the 2018 Conference of the North {A}merican
  Chapter of the Association for Computational Linguistics: Human Language
  Technologies, Volume 2 (Short Papers)}, pages 753--757, New Orleans,
  Louisiana. Association for Computational Linguistics.

\bibitem[Goo et~al., 2018b]{goo2018slot}
Goo, C.-W., Gao, G., Hsu, Y.-K., Huo, C.-L., Chen, T.-C., Hsu, K.-W., and Chen,
  Y.-N. (2018b).
\newblock Slot-gated modeling for joint slot filling and intent prediction.
\newblock In {\em Proceedings of the 2018 Conference of the North American
  Chapter of the Association for Computational Linguistics: Human Language
  Technologies, Volume 2 (Short Papers)}, pages 753--757.

\bibitem[Goodfellow et~al., 2014]{goodfellow2014explaining}
Goodfellow, I.~J., Shlens, J., and Szegedy, C. (2014).
\newblock Explaining and harnessing adversarial examples.
\newblock {\em arXiv preprint arXiv:1412.6572}.

\bibitem[Gopalakrishnan et~al., 2020]{gopalakrishnan2020neural}
Gopalakrishnan, K., Hedayatnia, B., Wang, L., Liu, Y., and Hakkani-Tur, D.
  (2020).
\newblock Are neural open-domain dialog systems robust to speech recognition
  errors in the dialog history? an empirical study.
\newblock {\em arXiv preprint arXiv:2008.07683}.

\bibitem[Gui et~al., 2021]{gui2021textflint}
Gui, T., Wang, X., Zhang, Q., Liu, Q., Zou, Y., Zhou, X., Zheng, R., Zhang, C.,
  Wu, Q., Ye, J., Pang, Z., Zhang, Y., Li, Z., Ma, R., Fei, Z., Cai, R., Zhao,
  J., Hu, X., Yan, Z., Tan, Y., Hu, Y., Bian, Q., Liu, Z., Zhu, B., Qin, S.,
  Xing, X., Fu, J., Zhang, Y., Peng, M., Zheng, X., Zhou, Y., Wei, Z., Qiu, X.,
  and Huang, X. (2021).
\newblock Textflint: Unified multilingual robustness evaluation toolkit for
  natural language processing.

\bibitem[Guo et~al., 2021]{guo2021learning}
Guo, Y., Shou, L., Pei, J., Gong, M., Xu, M., Wu, Z., and Jiang, D. (2021).
\newblock Learning from multiple noisy augmented data sets for better
  cross-lingual spoken language understanding.
\newblock {\em arXiv preprint arXiv:2109.01583}.

\bibitem[Gusfield, 1997]{Gusfield:97}
Gusfield, D. (1997).
\newblock {\em Algorithms on Strings, Trees and Sequences}.
\newblock Cambridge University Press, Cambridge, UK.

\bibitem[He et~al., 2020a]{he-etal-2020-syntactic}
He, K., Lei, S., Yang, Y., Jiang, H., and Wang, Z. (2020a).
\newblock Syntactic graph convolutional network for spoken language
  understanding.
\newblock In {\em Proceedings of the 28th International Conference on
  Computational Linguistics}, pages 2728--2738, Barcelona, Spain (Online).
  International Committee on Computational Linguistics.

\bibitem[He et~al., 2020b]{he2020multi}
He, K., Xu, W., and Yan, Y. (2020b).
\newblock Multi-level cross-lingual transfer learning with language shared and
  specific knowledge for spoken language understanding.
\newblock {\em IEEE Access}, 8:29407--29416.

\bibitem[He et~al., 2020c]{he-etal-2020-learning-tag}
He, K., Yan, Y., and Xu, W. (2020c).
\newblock Learning to tag {OOV} tokens by integrating contextual representation
  and background knowledge.
\newblock In {\em Proceedings of the 58th Annual Meeting of the Association for
  Computational Linguistics}, pages 619--624, Online. Association for
  Computational Linguistics.

\bibitem[He et~al., 2020d]{he2020learning}
He, K., Yan, Y., and Xu, W. (2020d).
\newblock Learning to tag oov tokens by integrating contextual representation
  and background knowledge.
\newblock In {\em Proceedings of the 58th Annual Meeting of the Association for
  Computational Linguistics}, pages 619--624.

\bibitem[Heigold et~al., 2018]{heigold-etal-2018-robust}
Heigold, G., Varanasi, S., Neumann, G., and van Genabith, J. (2018).
\newblock How robust are character-based word embeddings in tagging and {MT}
  against wrod scramlbing or randdm nouse?
\newblock In {\em Proceedings of the 13th Conference of the Association for
  Machine Translation in the {A}mericas (Volume 1: Research Track)}, pages
  68--80, Boston, MA. Association for Machine Translation in the Americas.

\bibitem[Hinton et~al., 2012]{hinton2012improving}
Hinton, G.~E., Srivastava, N., Krizhevsky, A., Sutskever, I., and
  Salakhutdinov, R.~R. (2012).
\newblock Improving neural networks by preventing co-adaptation of feature
  detectors.
\newblock {\em arXiv preprint arXiv:1207.0580}.

\bibitem[Hou et~al., 2020]{hou2020c2cgenda}
Hou, Y., Chen, S., Che, W., Chen, C., and Liu, T. (2020).
\newblock C2c-genda: Cluster-to-cluster generation for data augmentation of
  slot filling.

\bibitem[Hou et~al., 2018]{hou2018sequence}
Hou, Y., Liu, Y., Che, W., and Liu, T. (2018).
\newblock Sequence-to-sequence data augmentation for dialogue language
  understanding.
\newblock {\em arXiv preprint arXiv:1807.01554}.

\bibitem[Huang and Chen, 2020]{huang2020learning}
Huang, C.-W. and Chen, Y.-N. (2020).
\newblock Learning asr-robust contextualized embeddings for spoken language
  understanding.
\newblock In {\em ICASSP 2020-2020 IEEE International Conference on Acoustics,
  Speech and Signal Processing (ICASSP)}, pages 8009--8013. IEEE.

\bibitem[Krizhevsky et~al., 2017]{10.1145/3065386}
Krizhevsky, A., Sutskever, I., and Hinton, G.~E. (2017).
\newblock Imagenet classification with deep convolutional neural networks.
\newblock {\em Commun. ACM}, 60(6):84–90.

\bibitem[Krone et~al., 2021a]{krone2021robustness}
Krone, J., Sengupta, S., and Mansoor, S. (2021a).
\newblock On the robustness of goal oriented dialogue systems to real-world
  noise.
\newblock {\em arXiv preprint arXiv:2104.07149}.

\bibitem[Krone et~al., 2021b]{DBLP:journals/corr/abs-2104-07149}
Krone, J., Sengupta, S., and Mansoor, S. (2021b).
\newblock On the robustness of goal oriented dialogue systems to real-world
  noise.
\newblock {\em CoRR}, abs/2104.07149.

\bibitem[Lample et~al., 2016]{lample2016neural}
Lample, G., Ballesteros, M., Subramanian, S., Kawakami, K., and Dyer, C.
  (2016).
\newblock Neural architectures for named entity recognition.
\newblock {\em arXiv preprint arXiv:1603.01360}.

\bibitem[Le et~al., 2022]{le2022perturbations}
Le, T., Lee, J., Yen, K., Hu, Y., and Lee, D. (2022).
\newblock Perturbations in the wild: Leveraging human-written text
  perturbations for realistic adversarial attack and defense.
\newblock {\em arXiv preprint arXiv:2203.10346}.

\bibitem[Lewis et~al., 2019]{lewis2019bart}
Lewis, M., Liu, Y., Goyal, N., Ghazvininejad, M., Mohamed, A., Levy, O.,
  Stoyanov, V., and Zettlemoyer, L. (2019).
\newblock Bart: Denoising sequence-to-sequence pre-training for natural
  language generation, translation, and comprehension.
\newblock {\em arXiv preprint arXiv:1910.13461}.

\bibitem[Li et~al., 2020a]{li2020conditional}
Li, K., Chen, C., Quan, X., Ling, Q., and Song, Y. (2020a).
\newblock Conditional augmentation for aspect term extraction via masked
  sequence-to-sequence generation.
\newblock {\em arXiv preprint arXiv:2004.14769}.

\bibitem[Li et~al., 2020b]{li2020multi}
Li, M., Liu, X., Ruan, W., Soldaini, L., Hamza, W., and Su, C. (2020b).
\newblock Multi-task learning of spoken language understanding by integrating
  n-best hypotheses with hierarchical attention.
\newblock In {\em Proceedings of the 28th International Conference on
  Computational Linguistics: Industry Track}, pages 113--123.

\bibitem[Li et~al., 2020c]{li2020improving}
Li, M., Ruan, W., Liu, X., Soldaini, L., Hamza, W., and Su, C. (2020c).
\newblock Improving spoken language understanding by exploiting asr n-best
  hypotheses.
\newblock {\em arXiv preprint arXiv:2001.05284}.

\bibitem[Lin et~al., 2021]{lin2021rockner}
Lin, B.~Y., Gao, W., Yan, J., Moreno, R., and Ren, X. (2021).
\newblock Rockner: A simple method to create adversarial examples for
  evaluating the robustness of named entity recognition models.
\newblock {\em arXiv preprint arXiv:2109.05620}.

\bibitem[Lin et~al., 2020]{lin-etal-2020-rigorous}
Lin, H., Lu, Y., Tang, J., Han, X., Sun, L., Wei, Z., and Yuan, N.~J. (2020).
\newblock A rigorous study on named entity recognition: Can fine-tuning
  pretrained model lead to the promised land?
\newblock In {\em Proceedings of the 2020 Conference on Empirical Methods in
  Natural Language Processing (EMNLP)}, pages 7291--7300, Online. Association
  for Computational Linguistics.

\bibitem[Liu and Lane, 2015]{liu2015recurrent}
Liu, B. and Lane, I. (2015).
\newblock Recurrent neural network structured output prediction for spoken
  language understanding.
\newblock In {\em Proc. NIPS Workshop on Machine Learning for Spoken Language
  Understanding and Interactions}.

\bibitem[Liu and Lane, 2016a]{liu-lane-2016-joint}
Liu, B. and Lane, I. (2016a).
\newblock Joint online spoken language understanding and language modeling with
  recurrent neural networks.
\newblock In {\em Proceedings of the 17th Annual Meeting of the Special
  Interest Group on Discourse and Dialogue}, pages 22--30, Los Angeles.
  Association for Computational Linguistics.

\bibitem[Liu and Lane, 2016b]{Liu2016AttentionBasedRN}
Liu, B. and Lane, I.~R. (2016b).
\newblock Attention-based recurrent neural network models for joint intent
  detection and slot filling.
\newblock In {\em INTERSPEECH}.

\bibitem[Liu et~al., 2020]{liu2020robustness}
Liu, J., Takanobu, R., Wen, J., Wan, D., Li, H., Nie, W., Li, C., Peng, W., and
  Huang, M. (2020).
\newblock Robustness testing of language understanding in task-oriented dialog.
\newblock {\em arXiv preprint arXiv:2012.15262}.

\bibitem[Lu et~al., 2018]{lu2018visual}
Lu, D., Neves, L., Carvalho, V., Zhang, N., and Ji, H. (2018).
\newblock Visual attention model for name tagging in multimodal social media.
\newblock In {\em Proceedings of the 56th Annual Meeting of the Association for
  Computational Linguistics (Volume 1: Long Papers)}, pages 1990--1999.

\bibitem[Lu et~al., 2021]{lu-etal-2021-slot}
Lu, H., Han, Z., Yuan, C., Wang, X., Lei, S., Jiang, H., and Wu, W. (2021).
\newblock Slot transferability for cross-domain slot filling.
\newblock In {\em Findings of the Association for Computational Linguistics:
  ACL-IJCNLP 2021}, pages 4970--4979, Online. Association for Computational
  Linguistics.

\bibitem[Miller, 1995]{miller1995wordnet}
Miller, G.~A. (1995).
\newblock Wordnet: a lexical database for english.
\newblock {\em Communications of the ACM}, 38(11):39--41.

\bibitem[Moradi and Samwald, 2021a]{moradi2021evaluating}
Moradi, M. and Samwald, M. (2021a).
\newblock Evaluating the robustness of neural language models to input
  perturbations.
\newblock {\em arXiv preprint arXiv:2108.12237}.

\bibitem[Moradi and Samwald, 2021b]{eval2021}
Moradi, M. and Samwald, M. (2021b).
\newblock Evaluating the robustness of neural language models to input
  perturbations.
\newblock {\em Proceedings of the 2021 Conference on Empirical Methods in
  Natural Language Processing}.

\bibitem[Muralidharan et~al., 2020]{muralidharan2020noise}
Muralidharan, D., Moniz, J. R.~A., Gao, S., Yang, X., Kao, J., Pulman, S.,
  Kothari, A., Shen, R., Pan, Y., Kaul, V., et~al. (2020).
\newblock Noise robust named entity understanding for voice assistants.
\newblock {\em arXiv preprint arXiv:2005.14408}.

\bibitem[Namysl et~al., 2020]{namysl2020nat}
Namysl, M., Behnke, S., and K{\"o}hler, J. (2020).
\newblock Nat: noise-aware training for robust neural sequence labeling.
\newblock {\em arXiv preprint arXiv:2005.07162}.

\bibitem[Namysl et~al., 2021a]{namysl2021empirical}
Namysl, M., Behnke, S., and K{\"o}hler, J. (2021a).
\newblock Empirical error modeling improves robustness of noisy neural sequence
  labeling.
\newblock {\em arXiv preprint arXiv:2105.11872}.

\bibitem[Namysl et~al., 2021b]{nat2021}
Namysl, M., Behnke, S., and Köhler, J. (2021b).
\newblock Empirical error modeling improves robustness of noisy neural sequence
  labeling.
\newblock {\em Findings of the Association for Computational Linguistics:
  ACL-IJCNLP 2021}.

\bibitem[Niu et~al., 2019]{niu2019novel}
Niu, P., Chen, Z., Song, M., et~al. (2019).
\newblock A novel bi-directional interrelated model for joint intent detection
  and slot filling.
\newblock {\em arXiv preprint arXiv:1907.00390}.

\bibitem[Paszke et~al., 2019]{paszke2019pytorch}
Paszke, A., Gross, S., Massa, F., Lerer, A., Bradbury, J., Chanan, G., Killeen,
  T., Lin, Z., Gimelshein, N., Antiga, L., et~al. (2019).
\newblock Pytorch: An imperative style, high-performance deep learning library.
\newblock {\em Advances in neural information processing systems},
  32:8026--8037.

\bibitem[Peng et~al., 2021]{peng2021soloist}
Peng, B., Li, C., Li, J., Shayandeh, S., Liden, L., and Gao, J. (2021).
\newblock Soloist: Buildingtask bots at scale with transfer learning and
  machine teaching.
\newblock {\em Transactions of the Association for Computational Linguistics},
  9:807--824.

\bibitem[Peng et~al., 2020a]{peng2020raddle}
Peng, B., Li, C., Zhang, Z., Zhu, C., Li, J., and Gao, J. (2020a).
\newblock Raddle: An evaluation benchmark and analysis platform for robust
  task-oriented dialog systems.
\newblock {\em arXiv preprint arXiv:2012.14666}.

\bibitem[Peng et~al., 2020b]{peng2020data}
Peng, B., Zhu, C., Zeng, M., and Gao, J. (2020b).
\newblock Data augmentation for spoken language understanding via pretrained
  models.
\newblock {\em arXiv e-prints}, pages arXiv--2004.

\bibitem[Rasooli and Tetreault, 2015]{rasooli-tetrault-2015}
Rasooli, M.~S. and Tetreault, J.~R. (2015).
\newblock Yara parser: {A} fast and accurate dependency parser.
\newblock {\em Computing Research Repository}, arXiv:1503.06733.
\newblock version 2.

\bibitem[Ribeiro et~al., 2020a]{check2020}
Ribeiro, M.~T., Wu, T., Guestrin, C., and Singh, S. (2020a).
\newblock Beyond accuracy: Behavioral testing of nlp models with checklist.
\newblock {\em Proceedings of the 58th Annual Meeting of the Association for
  Computational Linguistics}.

\bibitem[Ribeiro et~al., 2020b]{ribeiro2020beyond}
Ribeiro, M.~T., Wu, T., Guestrin, C., and Singh, S. (2020b).
\newblock Beyond accuracy: Behavioral testing of nlp models with checklist.
\newblock {\em arXiv preprint arXiv:2005.04118}.

\bibitem[Rozsa et~al., 2016]{rozsa2016adversarial}
Rozsa, A., Rudd, E.~M., and Boult, T.~E. (2016).
\newblock Adversarial diversity and hard positive generation.
\newblock In {\em Proceedings of the IEEE Conference on Computer Vision and
  Pattern Recognition Workshops}, pages 25--32.

\bibitem[Ruan et~al., 2020]{ruan2020towards}
Ruan, W., Nechaev, Y., Chen, L., Su, C., and Kiss, I. (2020).
\newblock Towards an asr error robust spoken language understanding system.
\newblock In {\em INTERSPEECH}, pages 901--905.

\bibitem[Sergio et~al., 2020]{sergio2020attentively}
Sergio, G.~C., Moirangthem, D.~S., and Lee, M. (2020).
\newblock Attentively embracing noise for robust latent representation in bert.
\newblock In {\em Proceedings of the 28th International Conference on
  Computational Linguistics}, pages 3479--3491.

\bibitem[Shen et~al., 2020]{shen2020simple}
Shen, D., Zheng, M., Shen, Y., Qu, Y., and Chen, W. (2020).
\newblock A simple but tough-to-beat data augmentation approach for natural
  language understanding and generation.

\bibitem[Sperber et~al., 2017]{sperber2017toward}
Sperber, M., Niehues, J., and Waibel, A. (2017).
\newblock Toward robust neural machine translation for noisy input sequences.
\newblock In {\em International Workshop on Spoken Language Translation
  (IWSLT)}, page~18.

\bibitem[Sundararaman et~al., 2021]{sundararaman2021phoneme}
Sundararaman, M.~N., Kumar, A., and Vepa, J. (2021).
\newblock Phoneme-bert: Joint language modelling of phoneme sequence and asr
  transcript.
\newblock {\em arXiv preprint arXiv:2102.00804}.

\bibitem[Wang et~al., 2022a]{Wang2022InstructionNERAM}
Wang, L., Li, R., Yan, Y., Yan, Y., Wang, S., Wu, W.~Y., and Xu, W. (2022a).
\newblock Instructionner: A multi-task instruction-based generative framework
  for few-shot ner.
\newblock {\em ArXiv}, abs/2203.03903.

\bibitem[Wang et~al., 2021a]{wang-etal-2021-bridge}
Wang, L., Li, X., Liu, J., He, K., Yan, Y., and Xu, W. (2021a).
\newblock Bridge to target domain by prototypical contrastive learning and
  label confusion: Re-explore zero-shot learning for slot filling.
\newblock In {\em Proceedings of the 2021 Conference on Empirical Methods in
  Natural Language Processing}, pages 9474--9480, Online and Punta Cana,
  Dominican Republic. Association for Computational Linguistics.

\bibitem[Wang et~al., 2022b]{wang2022miner}
Wang, X., Dou, S., Xiong, L., Zou, Y., Zhang, Q., Gui, T., Qiao, L., Cheng, Z.,
  and Huang, X. (2022b).
\newblock Miner: Improving out-of-vocabulary named entity recognition from an
  information theoretic perspective.
\newblock {\em arXiv preprint arXiv:2204.04391}.

\bibitem[Wang et~al., 2021b]{wang2021randomized}
Wang, X., Xiong, Y., and He, K. (2021b).
\newblock Randomized substitution and vote for textual adversarial example
  detection.
\newblock {\em arXiv preprint arXiv:2109.05698}.

\bibitem[Wei and Zou, 2019]{wei2019eda}
Wei, J. and Zou, K. (2019).
\newblock Eda: Easy data augmentation techniques for boosting performance on
  text classification tasks.
\newblock {\em arXiv preprint arXiv:1901.11196}.

\bibitem[Weng et~al., 2020]{weng2020joint}
Weng, Y., Miryala, S.~S., Khatri, C., Wang, R., Zheng, H., Molino, P.,
  Namazifar, M., Papangelis, A., Williams, H., Bell, F., et~al. (2020).
\newblock Joint contextual modeling for asr correction and language
  understanding.
\newblock In {\em ICASSP 2020-2020 IEEE International Conference on Acoustics,
  Speech and Signal Processing (ICASSP)}, pages 6349--6353. IEEE.

\bibitem[Wu et~al., 2021]{wu2021bridging}
Wu, D., Chen, Y., Ding, L., and Tao, D. (2021).
\newblock Bridging the gap between clean data training and real-world inference
  for spoken language understanding.
\newblock {\em arXiv preprint arXiv:2104.06393}.

\bibitem[Wu et~al., 2022]{wu2022text}
Wu, X., Gao, C., Lin, M., Zang, L., Wang, Z., and Hu, S. (2022).
\newblock Text smoothing: Enhance various data augmentation methods on text
  classification tasks.
\newblock {\em arXiv preprint arXiv:2202.13840}.

\bibitem[Yan et~al., 2020]{yan-etal-2020-adversarial}
Yan, Y., He, K., Xu, H., Liu, S., Meng, F., Hu, M., and Xu, W. (2020).
\newblock Adversarial semantic decoupling for recognizing open-vocabulary
  slots.
\newblock In {\em Proceedings of the 2020 Conference on Empirical Methods in
  Natural Language Processing (EMNLP)}, pages 6070--6075, Online. Association
  for Computational Linguistics.

\bibitem[Yan et~al., 2021]{yan2021consert}
Yan, Y., Li, R., Wang, S., Zhang, F., Wu, W., and Xu, W. (2021).
\newblock Consert: A contrastive framework for self-supervised sentence
  representation transfer.

\bibitem[Yang et~al., 2022]{yang2022prompting}
Yang, Y., Huang, P., Cao, J., Li, J., Lin, Y., Dong, J.~S., Ma, F., and Zhang,
  J. (2022).
\newblock A prompting-based approach for adversarial example generation and
  robustness enhancement.
\newblock {\em arXiv preprint arXiv:2203.10714}.

\bibitem[Ye et~al., 2022]{ye2022assist}
Ye, F., Feng, Y., and Yilmaz, E. (2022).
\newblock Assist: Towards label noise-robust dialogue state tracking.
\newblock {\em arXiv preprint arXiv:2202.13024}.

\bibitem[Zhang et~al., 2018]{zhang2018adaptive}
Zhang, Q., Fu, J., Liu, X., and Huang, X. (2018).
\newblock Adaptive co-attention network for named entity recognition in tweets.
\newblock In {\em Thirty-Second AAAI Conference on Artificial Intelligence}.

\bibitem[Zhou et~al., 2021]{zhou2021melm}
Zhou, R., He, R., Li, X., Bing, L., Cambria, E., Si, L., and Miao, C. (2021).
\newblock Melm: Data augmentation with masked entity language modeling for
  cross-lingual ner.
\newblock {\em arXiv preprint arXiv:2108.13655}.

\end{thebibliography}
\bibliographystyle{acl_natbib}




\section{Baselines}
\label{ref:D}

To simulate the input perturbation existing in realistic scenarios, we introduce five well-designed perturbation robust methods and a strong baseline:

\textbf{Random Char Augmentation (Char-Random)} is a character-level augmentation method that randomly adds, removes, and replaces characters in a token with a transformation probability ${p}$.

\textbf{Random Word Deletion (Word-Del)} aims to simulate the effect of simplification in input utterances in real-world scenarios \cite{wei2019eda}. It randomly removes tokens with a probability ${p}$.

\textbf{Random Word Insertion (Word-Insert)} randomly insert words with probability ${p}$ based on contextual embedding \cite{peng2020data}. The method aims to model the effect of verbosity perturbation in input utterances.

\textbf{Homophonic substitution (Hom-Sub)} is designed for simulating word-level perturbation. We implement a homophone replacement dictionary, where words in the utterance are replaced by homophones with probability ${p}$.

\textbf{Synonymous Substitution (Syn-Sub)}   is implemented based on WordNet’s \cite{miller1995wordnet} synonymous thesaurus. We randomly select tokens in utterance with probability ${p}$ for synonymous substitution \cite{coulombe2018text}. Note that our augmentations on training samples avoid slot words and only operate on contextual words.

\textbf{Noise-Aware Training} is proposed by \cite{namysl2020nat}, which includes two Noise-Aware
Training (NAT) objectives that improve robustness of sequence labeling performed on perturbed input. The data augmentation method trains a neural model using a mixture of clean
and noisy samples, whereas the stability training algorithm encourages the model to create
a noise-invariant latent representation.

\section{BERT Result on RADDLE}

Table \ref{tab:noise-sf-bert} shows the BERT-version results of PSSAT. Compared to several data augmentation methods, PSSAT method makes a great improvement in each field. The overall results are better than any type of data augmentation results. Furthermore, the whole PSSAT method outperforms the baseline by 18.8\%. Similar to the results of LSTM, PSSAT also achieves the best results on each spoken language perturbation.


%



\section{SNIPS Single Perturbation Experiment}
\label{ref:F}
As shown in Table \ref{tab:snips single}, we also explore the performance of various denoising methods on SNIPS dataset. Both entity mention and contextual semantics are corrupted in mixed multiple noise scenarios, resulting in a catastrophic degradation of model performance. The overall result combining the single-noise and multi-noise results achieves a 33\% improvement.

\section{Case Study}
\label{ref:H}

Table \ref{tab:case}  shows some samples generated by PSSAT in the way of RWM and RCM, respectively. It can be seen that the generated augmented samples are more in line with the Spoken language perturbation, while preserving the semantics of the original sentences.

\begin{table*}[t]
  \centering
  \resizebox{0.9\textwidth}{!}{
  \begin{tabular}{lcccccccccc}
    \toprule
    \textbf{Methods}&\textbf{Clean}&\textbf{Homophone}&\textbf{paraphrase}&\textbf{verbose}&\textbf{simplification}&\textbf{Overall} \\
    
    \midrule
    none &
    96.2&
    82.8(-13.4)&
    90.4(-5.8)&
    84.4(-11.8)&
    87.7(-8.5)&
    82.6(-13.6)\\
    \midrule
    
    Char-Random &
    96.0 (-0.2)&
    85.0 (16.4\%)&
    \textcolor{red}{89.9 (-8.6\%)}&
    84.9 (4.2\%)&
    88.1 (4.7\%)&
    86.9 (4.2\%)\\
    
    Word-Del &
    95.9 (-0.3)&
    84.5 (12.7\%)&
    90.0 (-6.9\%)&
    84.5 (0.8\%)&
    88.0 (3.5\%)&
    86.8 (2.5\%)\\
    
    Word-Sub &
    96.3 (0.1)&
    84.1 (9.7\%)&
    90.2 (-3.4\%)&
    84.1 (-2.5\%)&
    88.2 (5.9\%)&
    86.7 (2.4\%)\\
    
    Word-Insert &
    96.3 (0.1)&
    84.3 (9.7\%)&
    90.5 (1.7\%)&
    83.9 (-4.2\%)&
    88.5 (9.4\%)&
    86.8 (4.2\%)\\    
    
    Homophone &
    95.8 (-0.4)&
    \textbf{85.8 (22.4\%)}&
    90.2 (-3.4\%)&
    \textcolor{red}{82.4 (-16.9\%)}&
    \textcolor{red}{87.5 (-2.4\%)}&
    \textcolor{red}{86.5 (-0.1\%)}\\   
    NAT($\mathcal{L}_{\textit{aug}}$) &
    96.0 (0.2)&
    85.2 (17.7\%)&
    90.5 (2.4\%)&
    85.4 (8.3\%)&
    88.0 (3.0\%)&
    87.2 (7.9\%)\\

    NAT($\mathcal{L}_{\textit{stabil}}$) &
    96.0 (0.2)&
    85.1 (16.8\%)&
    90.3(-1.2\%)&
    85.2 (6.6\%)&
    88.0 (3.0\%)&
    87.2 (6.3\%)\\  
    
     \midrule
    
    
    PSSAT  &
    96.4 (0.2)&
   \underline{ 85.6 (20.9\%)}&
    \textbf{91.5(19.0\%)}&
   \textbf{ 85.8(11.9\%)}&
    \textbf{89.7(23.5\%)}&
    \textbf{88.1(18.8\%)}\\
    
    \quad$-$ RCM &
    96.6 (0.4)&
    84.7 (14.0\%)&
    91.3 (15.5\%)&
    85.1 (5.9\%)&
    88.4 (8.2\%)&
    87.4(10.9\%) \\

    \quad$-$ RWM&
    96.4 (0.2)&
    83.5(5.2\%)&
    \textbf{91.5(19.0\%)}&
    \underline{85.7(11.0\%)}&
    \underline{89.4(20.0\%)}&
    \underline{87.5(13.8\%)}\\

    \quad$-$ Pre-training&
    95.9 (0.1)&
    \textcolor{red}{83.1(2.2\%)}&
    90.7(4.9\%)&
    84.9(4.4\%)&
    88.2(5.6\%)&
    86.7(4.3\%)\\
    \bottomrule
  \end{tabular}}
  \caption{The performance (F1 score) of the PSSAT on RADDLE. For cells in \textit{Baseline} row and \textit{Clean test} column, the numbers in the parenthesis indicate the change of F1 score over the baseline (96.2), while for other cells, the numbers in the parenthesis indicate the perturbation recovery rate (${p_r}$). In Overall column, we calculate the average F1 and ${p_r}$ of the four Spoken language perturbations respectively.}
  \label{tab:noise-sf-bert}
  \vspace{-0.5cm}
  
\end{table*}


\begin{table*}[htbp]
    \centering
    \tiny
    \renewcommand{\arraystretch}{1.3} 
    \setlength{\tabcolsep}{3pt}
    \resizebox{0.7\textwidth}{!}{
    \begin{tabular}{cccccccccc}
        \hline

           \textbf{Method} & \textbf{Clean} & \textbf{Hom} &  \textbf{App}  & \textbf{Concat} & \textbf{Overall} \\
        
        \hline
        Baseline (LSTM) & 93.9 & 62.2(-31.7) & 71.2(-22.7)  & 85.0(-8.9) & 72.8(-21.1) \\
        
        Char-Random  & 93.7 & \textbf{75.8(42.9\%)} &  74.0(12.3\%)  & 85.2(2.2\%) & 78.3(19.1\%)\\        

        Word-Del & 93.8 & 61.6(-1.9\%) &  69.2(-8.8\%)  & 85.3(3.4\%) & 72.0(-2.4\%)\\   
        
        Word-Sub & 93.8 & 65.7(11.0\%) &  73.3 (9.3\%)  & 84.1 (-10.1\%) & 74.4(3.4\%)\\ 
        
        Word-Insert & 92.8 & 63.9 (5.4\%) &  80.5 (41.0\%)  & 82.1 (-32.6\%) & 75.5(4.6\%)\\ 

        Homephone & 93.7 &70.1 (24.9\%) & 72.8 (7.0\%)  & 86.4 (15.7\%) & 76.4(15.9\%)\\ 
        
        NAT($\mathcal{L}_{\textit{aug}}$) &
        93.6 &
        69.1 (21.8\%)&
        74.7 (15.3\%)&
        85.5 (5.5\%)&
        76.4 (14.2\%)\\          
        
        NAT($\mathcal{L}_{\textit{stabil}}$) &
        93.6 &
        68.4 (19.6\%)&
        74.3 (13.8\%)&
        85.4 (4.7\%)&
        76.0 (12.7\%)\\       
        
        \hline
        
        PSSAT & 94.14 & 71.5(29.3\%) & \textbf{82.7(50.7\%)} &  \textbf{86.7(19.1\%)} & \textbf{80.3(33.0\%)}\\    
        
        \hline
        
    \end{tabular}}
    \vspace{-0.2cm}
    \caption{The performance of the best baseline and PSSAT on mixed perturbations. Con, APP and Home stand for ConcatSent, Appendirr and Homophone, respectively. }
    \label{tab:snips single}
    \vspace{-0.3cm}
\end{table*}

\begin{table*}[htbp]
    \scriptsize
  \centering
    \begin{tabular}{c|l|l}
    \hline
          & \multicolumn{1}{c|}{Ori.} & \multicolumn{1}{c}{Aug.} \bigstrut\\
    \hline
    \multirow{4}[2]{*}{Text} & can you please check for a \textbf{turkish restaurant}  ? & \green{so can you show me some} \textbf{turkish restaurant} ? \bigstrut[t]\\
          & does it have \textbf{4 stars} ? &  \green{does that rated} \textbf{4 stars} ? \\
          & \textbf{5} {people for the train please .} & \textbf{5}  \green{tkts R  needed} please . \\
          & i 'll be leaving \textbf{kings lynn} after \textbf{13:15} . & i 'm gonna leave \textbf{kings lynn} \green{@} \textbf{13:15} . \bigstrut[b]\\
    \hline
    \multirow{3}[2]{*}{Word} & ok  , how about \textbf{scudamores punting company} then . & \green{@HeyandIfhey} , how about \textbf{scudamores punting company} then . \bigstrut[t]\\
          & how about a \textbf{museum} ? & how \green{is} a \textbf{museum} ? \\
          & i am looking for a \textbf{hotel} please . & i am \green{sorry } for a \textbf{hotel} please . \bigstrut[b]\\
    \hline
    \end{tabular}%
    \caption{Some raw data and the corresponding enhanced data.}
  \label{tab:case}%
\end{table*}%

\end{document}